\documentclass[sigconf]{acmart}

\usepackage{graphicx}
\usepackage{subcaption}

\AtBeginDocument{%
  }

\begin{document}

\title{ProcessPainter: Learn Painting Process from Sequence Data}

\author{Yiren Song}
\authornote{Both authors contributed equally to this research.}
\email{yiren@nus.edu.sg}
\affiliation{%
  \institution{Show Lab, National University of Singapore}
  \country{Singapore}
}

\author{Shijie Huang}
\authornotemark[1]
\email{huangshijie@u.nus.edu}
\affiliation{%
  \institution{National University of Singapore}
  \country{Singapore}
}

\author{Chen Yao}
\email{cyao10@u.nus.edu}
\affiliation{%
  \institution{National University of Singapore}
  \country{Singapore}
}

\author{Xiaojun Ye}
\email{xiaojunye81@gmail.com}
\affiliation{%
  \institution{ZheJiang University}
  \city{Hang Zhou}
  \state{Zhe Jiang}
  \country{China}}
    
\author{Hai Ci}
\email{hai.ci@nus.edu.sg}
\affiliation{%
  \institution{Show Lab, National University of Singapore}
  \country{Singapore}
}

\author{Jiaming Liu}
\email{jmliu1217@gmail.com}
\affiliation{%
  \institution{Tiamat}
  \city{Shanghai}
  \country{China}}

\author{Yuxuan Zhang}
\email{zyx153@sjtu.edu.cn}
\affiliation{%
  \institution{Shanghai Jiao Tong University}
  \city{Min hang}
  \city{Shanghai}
  \country{China}}
  
\author{Mike Zheng Shou}
\authornote{Corresponding author.}
\email{mike.zheng.shou@gmail.com}
\affiliation{%
  \institution{Show Lab, National University of Singapore}
  \country{Singapore}
}



\begin{abstract}
  The painting process of artists is inherently stepwise and varies significantly among different painters and styles. Generating detailed, step-by-step painting processes is essential for art education and research, yet remains largely underexplored. Traditional stroke-based rendering methods break down images into sequences of brushstrokes, yet they fall short of replicating the authentic processes of artists, with limitations confined to basic brushstroke modifications. Text-to-image models utilizing diffusion processes generate images through iterative denoising, also diverge substantially from artists' painting process. To address these challenges, we introduce ProcessPainter, a text-to-video model that is initially pre-trained on synthetic data and subsequently fine-tuned with a select set of artists' painting sequences using the LoRA model. This approach successfully generates painting processes from text prompts for the first time. Furthermore, we introduce an Artwork Replication Network capable of accepting arbitrary-frame input, which facilitates the controlled generation of painting processes, decomposing images into painting sequences, and completing semi-finished artworks. This paper offers new perspectives and tools for advancing art education and image generation technology. Our code is available at: \url{https://github.com/nicolaus-huang/ProcessPainter}.
\end{abstract}


\begin{teaserfigure}
  \includegraphics[width=\textwidth]{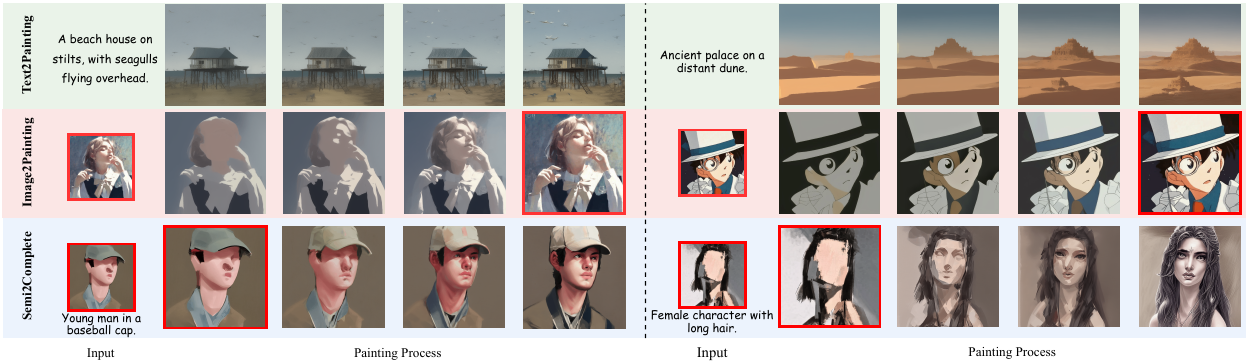}
  \caption{This paper accomplishes the generation of a painting process that mimics human artists, transitioning from abstract to specific. Each row in the figure represents a different application: (1) generating painting processes from text prompt (Text2Painting), (2) converting artworks into painting processes (Image2Painting), (3) completing semi-finished paintings (Semi2Complete).}
  \Description{.}
  \label{fig:teaser}
\end{teaserfigure}



\maketitle



\section{Introduction}




The step-by-step painting process has a wide range of applications, making it highly suitable for educational, entertainment, and professional fields. Generating a painting sequence that replicates the style of human artists is a long-desired but challenging goal\cite{hertzmann2003survey}. Previous studies have employed stroke-based rendering methods using stepwise greedy search\cite{vanderhaeghe2012stroke}, recurrent neural networks\cite{aksan2020cose,cao2019ai}, reinforcement learning\cite{schaldenbrand2021content}, and parameter searching\cite{tong2022im2oil,hu2023stroke} to simulate the human painting process. The Intelli-Paint\cite{singh2022intelli} utilizes semantic segmentation and hierarchical rendering to mimic the sequence of human painting. However, these methods have limited success in emulating the diverse painting processes of different artists, subjects, and art forms. Moreover, the process of human painting involves rich strategies and intelligence; painting subjects and backgrounds separately is just one of many strategies. These methods do not learn from human painting process key-frames and merely focus on minimizing similarity loss with input images by preventing brush strokes.

In the era of generative AI, the intersection of artificial intelligence and creative arts has seen significant advances in image synthesis technology, allowing the conversion of textual descriptions into visual representations\cite{zhou2021lafite,yang2023reco,cho2024visual}. Models typically based on diffusion\cite{li2024blip,ge2023preserve} processes have opened new horizons for artistic exploration, yet they still fail to generate step-by-step painting processes akin to those of human artists, making them unsuitable for creating step-by-step painting tutorials. Because painting is generally a gradual instantiation process, moving from abstract to specific, from macro to detail\cite{liu2021paint}, which is far removed from how diffusion models generate images through a denoising process. Decoding the latent from the denoising process through a VAE decoder only yields blurry images, not meaningful painting processes.

To address this, we introduce ProcessPainter, a data-driven solution that can generate vivid and realistic painting processes with controllable styles. ProcessPainter leverages temporal models trained on synthetic datasets and artists' painting sequences, defining the task of generating painting processes as a specific video generation problem. However, training or fine-tuning video models require massive amounts of data, and high-quality serialized painting video data are scarce. Videos documenting painting processes on the internet often include obstructions from the painter's hands, changes in camera angles, and zoom, making them unsuitable as training data. To innovate, we propose using an SBR-based method to construct synthetic datasets for training foundational painting models, enabling the generation of painting processes from textual descriptions. We then fine-tune the model with a small number of human painting sequences using LoRA\cite{hu2021LoRA} to learn the styles and process strategies of human artists. Painting LoRA not only replicates the artists' final visual styles but also their unique creative processes, providing a detailed observation of 'how' art is made, a feature largely missing in traditional diffusion models and SBR applications.

In painting education and practice, students copying classic works and teachers completing and modifying students' semi-finished pieces are two important teaching steps. To enable ProcessPainter as an AI tool to assist in these steps, we propose the Artwork Replication Network, which implements two practical functionalities: converting artworks to process key-frames and completing semi-finished paintings. Specifically, we trained an Artwork Replication Network that can accept any frame as a reference condition. During inference, inputting the reference image as the last frame enables generating a sequence from a blank canvas to the reference image. If the semi-finished reference image is input as a mid-sequence condition, ProcessPainter can progressively complete the reference image. We tested our method on various real-world images and paintings of different styles, and our experimental results demonstrate that our method can generate vivid painting processes with a high degree of anthropomorphism and artistic sense. Furthermore, our method is plug-and-play; by replacing different style Unets or mounting LoRA models, our approach can generate painting processes in various styles, such as oil painting, sketching, and ink painting.

The contributions of our paper are summarized as follows:
\begin{itemize}
\item We introduce a novel method for generating painting processes from textual descriptions, conceptualizing it as a video generation process.
\item We propose a method that involves pre-training the painting model on synthetic data, followed by training Painting LoRA to learn specific strategies and styles from a few human painting key-frames.
\item We propose an Artwork Replication Network that accommodates control through arbitrary-frame images, enabling the conversion of reference images into painting process and the completion of semi-finished paintings.
\end{itemize}

\section{Related Works}
\subsection{Stroke-based Rendering}


The problem of teaching machines “how to paint” has been extensively studied in the context of stroke-based rendering (SBR), which focuses on the recreation of non-photorealistic imagery through appropriate positioning and selection of discrete elements such as paint strokes or stipples\cite{hertzmann2003survey}. Early SBR approaches involved either a greedy, iterative search or required user interaction to address the decomposition problem \cite{haeberli1990paint,litwinowicz1997processing,hertzmann2022toward, 3Dstroke}. In recent years, recurrent neural networks (RNNs) and reinforcement learning (RL) have been extensively employed to generate strokes sequentially \cite{ha2017neural,zhou2018learning,xie2013artist,singh2022intelli}. Moreover, integrating adversarial training \cite{nakano2019neural} has emerged as a viable method for generating non-deterministic stroke sequences. adversarial updates of actors, critics, and discriminators. Stylized Neural Painting\cite{snp, kotovenko2021rethinking} proposed stroke optimization strategy that iteratively searches optimal parameters for each stroke and and is possible to be optimized jointly with neural style transfer. Similarly, the field of vector graphic generation also constructs images using a set of strokes and graphic elements\cite{clipdraw, clipvg, cliptexture, clipfont} . Although the aforementioned methods achieve the painting process from abstract to specific, these methods still significantly differ from the creative processes of human artists. This is because the painting processes of different artists and subjects vary significantly, making it challenging for a single method to simulate all. Recently, the emergence of Stable Diffusion\cite{rombach2022high} technology has significantly improved the effectiveness of generative images. Exploring the generation of painting processes based on diffusion models is a promising new field. To this end, we propose ProcessPainter, which fine-tunes diffusion models using data from artists' painting processes to learn their true distributions.

\subsection{Text2image Diffusion Model}
Recent studies have demonstrated that diffusion models are capable of generating high-quality synthetic images, effectively balancing diversity and fidelity. Models based on diffusion models or their variants, such as those documented in \cite{nichol2021glide, ramesh2022hierarchical, saharia2022photorealistic}, have successfully addressed the challenges associated with text-conditioned image synthesis. Stable Diffusion \cite{rombach2022high}, a model based on the Latent Diffusion Model, incorporates text conditioning within a UNet framework to facilitate text-based image generation, establishing itself as a mainstream model in image generation \cite{podell2023sdxl,li2023gligen,feng2023ranni}. Fine-tuning pre-trained image generation models can enhance their adaptation to specific application scenarios, as seen in techniques like LoRA \cite{hu2021LoRA} and DreamBooth \cite{ruiz2023dreambooth}. For theme control in text-to-image generation, several works \cite{kumari2023multi, ye2023ip, wang2024instantid, ssr, fast_icassp} focus on custom generation for defined pictorial concepts, with ControlNet \cite{zhang2023adding} additionally offering control over other modalities such as depth information. These Diffusion models have revolutionized the field of image generation with their powerful generative capabilities. However, the iterative denoising process is completely different from human painting, and cannot be mimicked with a brush, thus offering limited guidance for painting instruction. This paper attempts to use pre-trained diffusion models to generate painting processes that resemble those of artists.


\begin{figure*}[htbp!]
    \centering
    \includegraphics[width=1\linewidth]{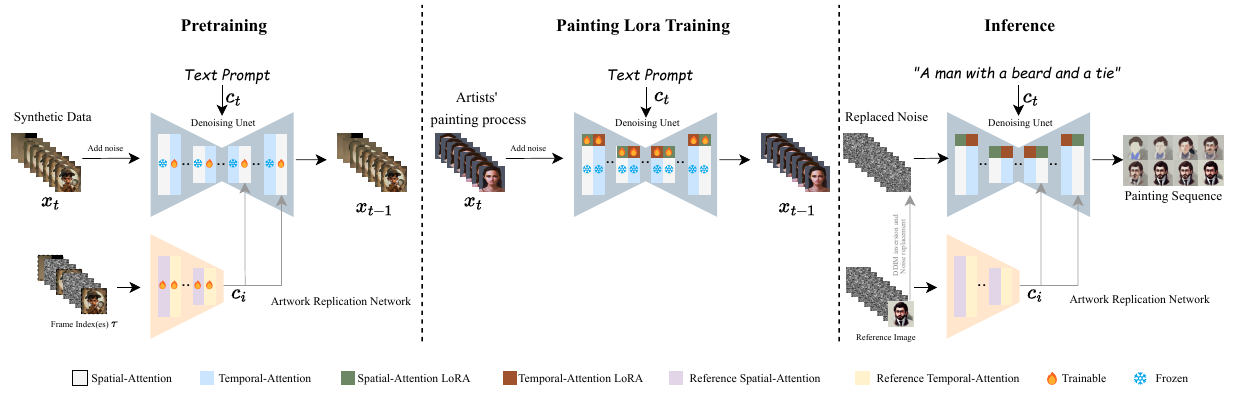}
    \caption{Overall schematics of our method. During the training phase, we first pre-train the Painting model and Artwork Replication Network on 40,000 synthetic data points. Then, we fine-tune the Painting LoRA model on a small amount of artists' painting process data. During inference, ProcessPainter generates the painting process step by step from a reference image, producing the final painting as the last frame. It can also refine a partially completed image based on textual descriptions and the input image. When no reference image is provided, ProcessPainter generates the painting process solely from textual descriptions.}
    \label{fig:architecture}
\end{figure*}

\subsection{Video Generation Models}

The foundational frameworks for image and video generation share similarities, primarily encompassing three pathways: Generative Adversarial Networks (GANs) \cite{tian2021good, shen2023mostgan}, autoregressive models \cite{le2021ccvs, ge2022long, hong2022cogvideo}, and diffusion models. Within this context, stable diffusion-based video generation models utilize training of 3D U-Nets to restore high-fidelity videos from Gaussian noises \cite{guo2023animatediff, lin2024animatediff, blattmann2023stable, khachatryan2023text2video}. To enhance the controllability of generated videos, recent advancements in motion customization video generation models employ multimodal control signals to influence the visual appearance and cinematographic effects of the outputs \cite{dai2023animateanything, ma2024follow, ma2024follow2}. Specifically, MotionCtrl \cite{wang2023motionctrl} facilitates precise control over camera and object motion, while ControlVideo inherits the architecture and weights from ControlNet \cite{zhang2023adding}, allowing for controllable video generation through motion sequences. Notably, there are no existing models for controllable video generation aimed at teaching painting processes. Thus, we propose ProcessPainter, which explores the creation of videos in various styles of painting processes, introducing a novel domain in stable diffusion-based video generation technology.

\section{Method}

In this section, we begin by exploring the preliminaries on diffusion models as
detailed in Section 3.1. Then introduce the Network architecture of our method inSection 3.2, followed by detailed descriptions of the key modules: the Painting Model, Painting LoRA, and Artwork Replication Network. In Section 3.3, we discuss the construction methods of the dataset and conclude with the settings for the training and inference phases in Section 3.4.

\subsection{Preliminary}
 


\textbf{Diffusion Model (DM).} DM \cite{ho2020denoising, wang2024stablegarment} belongs to the category of generative models that denoise from a Gaussian prior $x_T$ to target data distribution $x_0$ by means of an iterative denoising procedure. The common loss used in DM is:
\begin{equation}
L_{\text{simple}}(\theta) := \mathbb{E}_{x_0, t, \epsilon} \left[ \| \epsilon - \epsilon_\theta (x_t, t) \|^2_2 \right],
\end{equation}
where $x_t$ is a noisy image constructed by adding noise $\epsilon \sim \mathcal{N}(0, 1)$ to the natural image $x_0$ and the network $\epsilon_\theta(\cdot)$ is trained to predict the added noise. At inference time, data samples can be generated from Gaussian noise $\epsilon \sim \mathcal{N}(0, 1)$ using the predicted noise $\epsilon_\theta(x_t, t)$ at each timestep $t$ with samplers like DDPM \cite{ho2020denoising} or DDIM \cite{song2020denoising}.

\noindent \textbf{Latent Diffusion Model (LDM).} LDM \cite{rombach2022high} is proposed to model image representations in the autoencoder's latent space. LDM significantly speeds up the sampling process and facilitates text-to-image generation by incorporating additional text conditions. The LDM loss is:
\begin{equation}
L_{\text{LDM}}(\theta) := \mathbb{E}_{z_0, t, \epsilon} \left[ \| \epsilon - \epsilon_\theta (z_t, t, \tau_\theta(c_t)) \|^2_2 \right],
\end{equation}
where $z_0$ represents image latents and $\tau_\theta(\cdot)$ refers to the BERT text encoder \cite{devlin2018bert} used to encode text description $c_t$.

\noindent \textbf{Stable Diffusion (SD).} SD is a widely adopted text-to-image diffusion model based on LDM. Compared to LDM, SD is trained on the large LAION \cite{schuhmann2021laion} dataset and replaces BERT with the pre-trained CLIP \cite{radford2021learning} text encoder.

\subsection{Network Architecture}
\subsubsection{Overall Architecture.}

The overall architecture of ProcessPainter is depicted in Fig. 2. The
model is segmented into two primary components: a pre-trained stable diffusion model\cite{rombach2022high} with Temporal-Attention\cite{guo2023animatediff}, and an Artwork Replication Network. The parameters of the pre-trained stable diffusion, including the VAE and UNet, are fixed. We add a temporal module to each layer of the UNet, which is used to learn the inter-frame correlations of the progress key-frames. Additionally, an Artwork Replication Network accepts input from
the image context. The Artwork Replication Network is integrated with the denoising UNet in an additive fashion, thereby offering the flexibility for seamless model switching.



\subsubsection{Temporal-Painting Model and Painting LoRA} 




The key to generating painting sequences is that the entire sequence represents the transformation of the same image from abstract to detailed, ensuring consistency and correlation in content and composition across frames. To achieve this, we introduced the temporal attention module from AnimateDiff\cite{guo2023animatediff} into the UNet. 
This module, positioned after each diffusion layer, employs an inter-frame self-attention mechanism to assimilate information across different frames, ensuring smooth transitions and continuity throughout the sequence. 
This empirically validated training strategy grounds the model's capability in inter-frame self-attention, where dividing the reshaped feature map along the temporal axis results in vector sequences of length \( f \) (i.e., \(\{z_1, \ldots, z_f; z_i \in \mathbb{R}^{(b \times h \times w) \times c}\}\)), processed through self-attention blocks, i.e.
\begin{equation}
\text{Temporal-Attention}(Q, K, V) = \text{Softmax}\left(\frac{QK^T}{\sqrt{c}}\right) \cdot V,
\end{equation}
where $ Q = W^Q z $, $ K = W^K z $, and $ V = W^V z $ are separate linear projections. This attention mechanism allows the current frame's generation to utilize information from other frames.



Unlike video generation, the painting process undergoes more drastic changes from start to finish. The initial frame often consists of low-completion color blocks or sketches, while the final frame is a complete artwork. This presents a challenge for model training. To address this, we first pre-train the temporal module on a large synthetic dataset to learn the step-by-step painting processes using various Stroke- based-rendering methods. Then, we fine-tune the Painting LoRA model with 10-50 painting sequences from real artists. Notably, to decouple the artistic style of static works from the painting process strategies, we adopt a staged training approach for Painting LoRA. Initially, we train the Unet Spatial-Attention LoRA model using only the final frames of the painting sequences. Subsequently, we train the Temporal-Attention LoRA model on the full painting sequences. LoRA adds pairs of rank-decomposition matrices and optimizes only these newly introduced weights. By limiting the trainable parameters and keeping the original weights frozen, LoRA is less likely to cause catastrophic forgettingConcretely, the rank-decomposition matrices serve as the residual of the pre-trained model weights \( \mathcal{W} \in \mathbb{R}^{m \times n} \).
\begin{equation}
\mathcal{W}^{\prime}=\mathcal{W}+\Delta \mathcal{W}=\mathcal{W}+A B^T
\end{equation}

where \( \mathbf{A} \in \mathbb{R}^{m \times r} \), \( \mathbf{B} \in \mathbb{R}^{n \times r} \) are a pair of rank-decomposition matrices, \( r \) is a hyper-parameter, which is referred to as the rank of LoRA layers. In practice, LoRA is only applied to attention layers, further reducing the cost and storage for model fine-tuning.

\subsubsection{Artwork Replication Network.}

In the previous section, we explored generating painting processes from textual descriptions. In this section, we introduce the Artwork Replication Network, which accepts arbitrary-frame inputs to achieve controllable painting process generation. It offers two practical functionalities: converting reference images to process key-frames and completing semi-finished paintings.

Similar to previous controllable generation methods\cite{zhang2023controlvideo, zhang2023adding, sparse}, we introduce a variant of ControlNet to ensure that a specific frame \(\tau\) in the generated sequence matches the reference image. As shown in Figure \ref{fig:architecture}, during training and inference, the reference image is first encoded by the VAE encoder to obtain the latent representation \(l_i\). We initialize a standard Gaussian noise tensor as the input for the Artwork Replication Network and replace the standard Gaussian noise at the position corresponding to frame \(\tau\) with \(l_i\). We add a Temporal-Attention layer after each Spatial-Attention layer in ControlNet to learn how the reference image should influence the entire sequence. The final output of the n-th layer of the denoising UNet $\hat{S_n^u}$ can be define as:
\begin{equation}
\hat{S_n^u}=S_n^u + \lambda ARN_n
\end{equation}
where $S_n^u$ is the output feature of the denoising UNet. $\lambda ARN_n$ is the output feature of the Artwork Replication Network. \( n \) is the \( n \)-th layer in the Unet model. $\lambda$ is the coefficient for the degree of influence of the Artwork Replication Network.






During inference, The Artwork Replication Network supports multiple creative functions through different settings. When we set \(\tau\) to the final frame, the reference image input will serve as the reconstruction target for the last frame of the painting process, allowing us to convert reference images to process key-frames. Similarly, when we set \(\tau\) to the initial frames and input a partially completed painting as the reference, we achieve the effect of completing the reference image. The extent to which the network completes the partially finished painting depends on the setting of \(\tau\). The closer \(\tau\) is to the beginning, the greater the degree of painting completion.

\begin{figure*}[!htbp]
    \centering
    \includegraphics[width=1\linewidth]{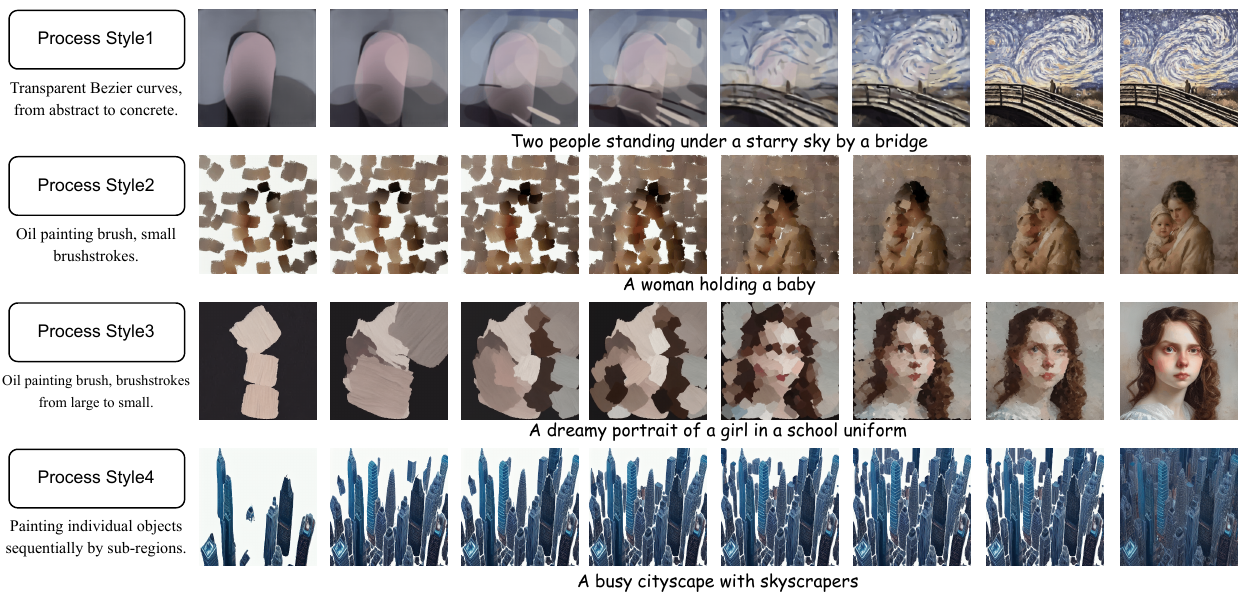}    \caption{Text to painting processe generation results. ProcessPainter can learn different process style from synthetic data. }
    \label{fig:text2painting}
\end{figure*}

\subsection{Dataset Construction Method}



Our dataset construction includes 30,000 synthetic painting sequences for pretraining, each with 8 frames at a resolution of 512x512, and 95 painting sequences from artists for LoRA trainting.

\noindent \textbf{Synthetic Data.}
We first selected 10,000 images from the DiffusionDB dataset based on aesthetic scores, then converted the static images into painting sequences using three SBR methods: Learn-to-Paint, Stylized Neural Painting, and Paint Transformer. Since the last few hundred steps of SBR methods mainly optimize details with minimal changes to the overall image, we used non-uniform sampling to select 8 frames based on each method's characteristics as the training set. Additionally, we proposed a method to simulate artists painting sub-regions from the foreground to the background for creating a synthetic dataset. Using Segment Anything\cite{kirillov2023segany} and Depth Anything\cite{depthanything}, we segmented all objects in the image and added them to a blank canvas in order from nearest to farthest based on depth. 

\noindent \textbf{Artists’ Data.}
Besides synthetic data, we also collected a total of 95 painting sequences from three artists. The painting types include impasto portraits, landscape sketches, and coloring line art. For each artist's work, we trained a separate Painting LoRA to learn their painting process strategies and styles. We add trigger word "sks" in the image captions to describe the painting process we wish to train on.

\subsection{Model Training and Inference}

\begin{figure*}[!htbp]
    \centering
    \includegraphics[width=1\linewidth]{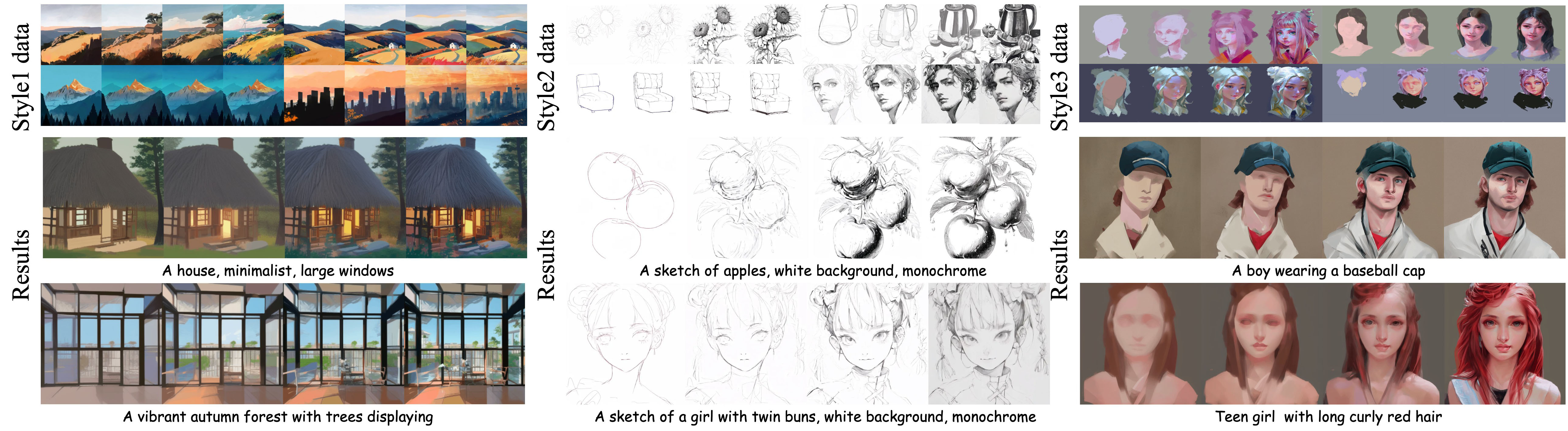}    \caption{
A Painting LoRA can be fine-tuned only on 10-50 sequences of artists' painting process, which can effectively capture the characteristics of the artists' painting process and the style of the final results. }
    \label{fig:lora}
\end{figure*}


\begin{figure*}[htbp!]
    \centering
    \includegraphics[width=1\linewidth]{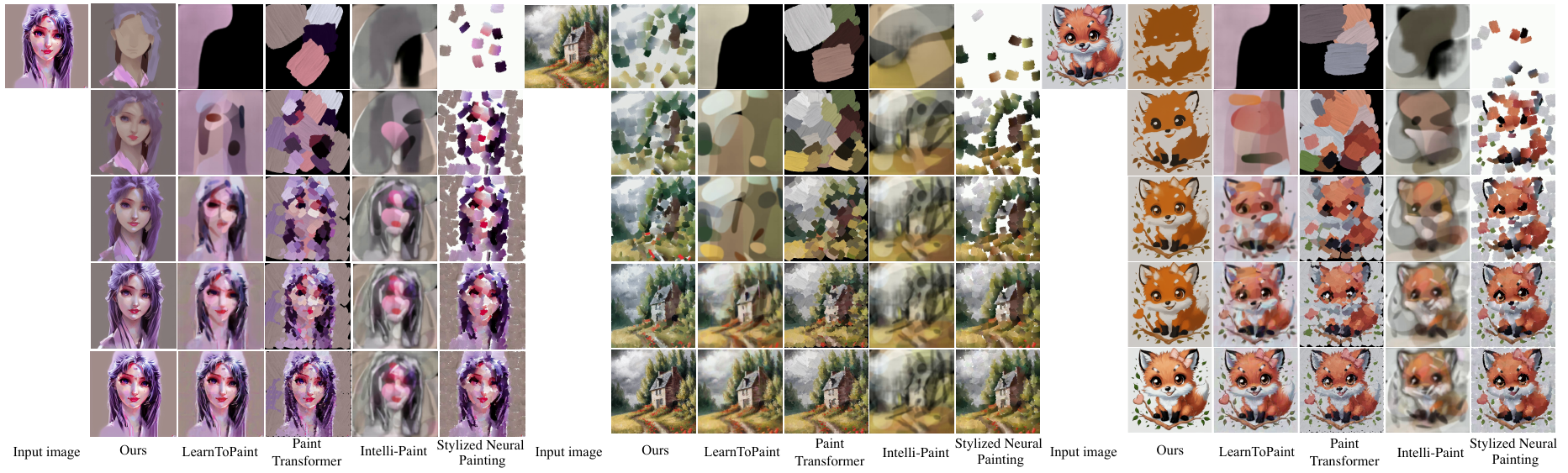}
    \caption{Compare with stroke based rendering methods, our method provides a more precise reconstruction of the original images, and the painting process more closely resembles that of human artists. For different types of paintings, the strategies of the painting process can be controlled by switching the Painting Model or Painting LoRA. }
    \label{fig:compare}
\end{figure*}

\begin{figure}[htb]
     \centering
     \begin{subfigure}{\linewidth}
         \centering
         \includegraphics[width=0.95\textwidth]{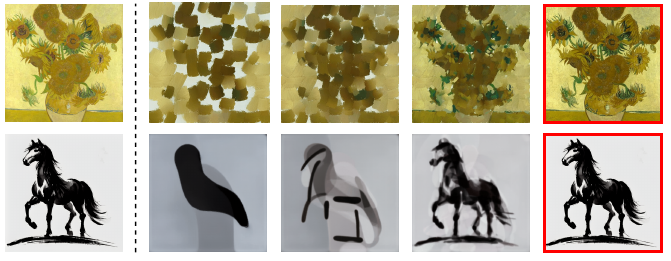}
         \caption{Converting artworks to painting process.}
         \label{fig:4_1}
    \end{subfigure}
    \hfill
     \begin{subfigure}{\linewidth}
         \centering
         \includegraphics[width=0.95\textwidth]{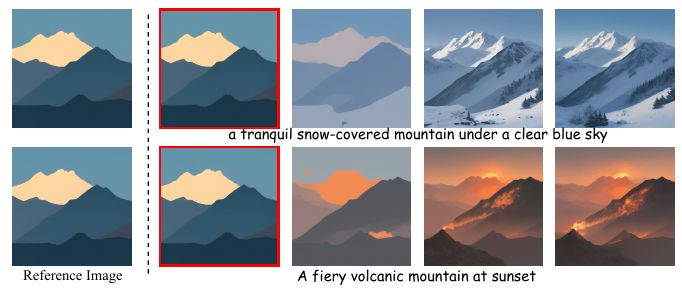}
         \caption{Completing semi-finished paintings.}
         \label{fig:4_2}
    \end{subfigure}
\caption{Given a reference image, ProcessPainter can generate a painting process that ensures specific frames match the reference image exactly.}
\vspace{-0.2cm}
\label{fig:ablation}
\end{figure}


In the training phase, we first pre-trained the Painting Model on a synthetic dataset. Then, we froze the parameters of the Painting Model and trained the Artwork Replication Network. For fine-tuning the Painting LoRA, we initially fine-tuned the Spatial-Attention LoRA using only the final frame to prevent unfinished painting works from damaging the image quality of the model. After this step, we froze the parameters of the Spatial-Attention LoRA and fine-tuned the Temporal-Attention LoRA using the complete painting sequence. We chose Prodigy as optimizer, which optimize the algorithm's traversal of the optimization landscape by dynamically adjusting the learning rate for each parameter based on their past gradients, can be especially beneficial for LoRA training.

During the inference phase, when generating painting sequences from text, we do not use the Artwork Replication Network. For tasks involving image-to-painting sequence conversion and painting completion, we use the Artwork Replication Network to receive the reference input for specific frames. To ensure that the frames in the generated painting sequence that should match the input image do so as closely as possible, we employ the DDIM Inversion technique to obtain the initial noise of the reference image and replace the initial noise in the UNet for the specific frames.

\section{Experiment}

\subsection{Experiment Setting}



We conducted a pre-training session using 30,000 video sequences over 50,000 steps, enabling the painting model to effectively capture the transition from textual semantics to painting processes. Subsequently, the model underwent 50,000 steps of formal training, employing a randomly sampled Artwork Replication Network. Our frame sampling strategy included extracting between 0 and 3 frames, with a sampling probability of one-third for both the first and last frames and a normal distribution for the middle frames, peaking at the center. This strategy was designed to mimic realistic usage. We optimized the frame extraction method to facilitate convergence and to align with actual probability distributions. The learning rate was set at \(2 \times 10^{-5}\), with a batch size of 1, at a resolution of 512*512 pixels, and the number of frames per sequence fixed at 8. Pre-training and LoRA training were conducted on a NVIDIA A100 GPU. During inference, we employed DDIM sampler and DDIM inversion techniques for noise replacement, running each for 50 steps.

\subsection{Text to Painting Process}

In this section, we demonstrate the capability of a foundational model trained with synthetic data to generate paintings from text descriptions. As shown in Figure \ref{fig:text2painting}, the first row illustrates the painting process from rough to detailed using semi-transparent Bezier curves as brush strokes. The second row displays the step-by-step creation of an oil painting on a blank canvas using oil paint brushes. The third row shows the painting process from the top left corner to the bottom right of the canvas. These painting patterns were learned from different painting process strategies synthesized from data. Figure \ref{fig:lora} displays the results of training the Painting LoRA model with a small number of key frames from human artists' processes, showcasing three implementations: color block landscape sketching, line drawing coloring, and impasto portraits. The experiments demonstrated that with just a dozen or so samples and based on pre-trained painting model, we can replicate the strategies and styles of painters.

\subsection{Image to Painting Process}
In this section, we present the experimental results based on the Artwork Replication Network. As shown in Fig. \ref{fig:4_1}, by controlling the last frame with a reference image, we have achieved the conversion of artworks into painting processes. In this task, a text prompt is not necessary. We can control different strategies of the painting process by swapping out the painting model or the painting LoRA model, for instance, using oil paint brush strokes to render Van Gogh's Starry Night or translucent strokes for Chinese ink paintings. When using a semi-finished piece as the reference image to control the first frame, we implemented text-guided painting completion capabilities. As displayed in Figure \ref{fig:4_2}, the results from ProcessPainter align well with the text prompts, demonstrating good consistency. Based on the same semi-finished image, different results can be obtained through various text prompts.

\subsection{Comparison and evaluation}
Since our method is the first to achieve text-to-painting-process generation and step-by-step painting completion, there are no existing methods for comparison. In this section, we compare our method with state-of-the-art baseline methods in the Image-to-Painting-Process task. We chose 4 stroke-based-rendering methods for Comparison, they are  LearnToPaint\cite{huang2019learning}, Paint Transformer\cite{liu2021paint}, Intelli-paint\cite{singh2022intelli}, and Stylized Neural Painting\cite{snp}.  

\subsubsection{Qualitative Evaluation}
As shown in Figure \ref{fig:compare}, although baseline methods can effectively replicate brushstroke processes, they are fundamentally designed to minimize the difference between the real image and the current canvas, resulting in a painting process that does not conform to human painting habits. Our method is based on stable diffusion, possesses a robust image prior, and has been fine-tuned with the painting processes of artists, thus aligning our results with the typical artist painting process. This approach showcases starting with broad color blocks and gradually adding details. Furthermore, the results from ProcessPainter exhibit more refined details and higher reconstruction accuracy relative to the reference image.


\begin{table}[h!]
\centering
\footnotesize
\caption{Quantitative Evaluation of Reconstruction Consistency}
\begin{tabular}{lccc}
\toprule
Method & Mean MSE $\downarrow$ & Mean LPIPS $\downarrow$ & Mean L1 $\downarrow$ \\
\midrule
LeranToPaint\cite{huang2019learning} & 0.016181 & 0.033240 & 0.087082 \\
Paint Transformer\cite{liu2021paint} & 0.087695 & 0.153372 & 0.187685 \\
Intelli-Paint\cite{singh2022intelli} & 0.247486 & 0.397536 & 0.350746 \\
Stylized Neural Painting\cite{snp} & 0.084447 & 0.141126 & 0.185832 \\
Ours & \textbf{0.014820} & \textbf{0.024517} & \textbf{0.082165} \\
\bottomrule
\end{tabular}
\label{tab1}
\end{table}

\subsubsection{Quantitative Evaluation}
In this section, we present the quantitative evaluation results. We evaluate the consistency between the output of the last frame in the image2painting task and the reference image. Specifically, we calculate the Mean Squared Error (MSE), Learned Perceptual Image Patch Similarity (LPIPS), and Least Absolute Deviations (L1). As a baseline comparison, we use LeranTopaint, Paint Transformer, Stylize neural painting, and Semantic RL, with results obtained from their official codebases and recommended parameters. Table~\ref{tab1} shows that our method achieves the highest reconstruction consistency compared with baseline methods.

\subsection{Ablation Study}
\begin{figure}
    \centering
    \includegraphics[width=0.9\linewidth]{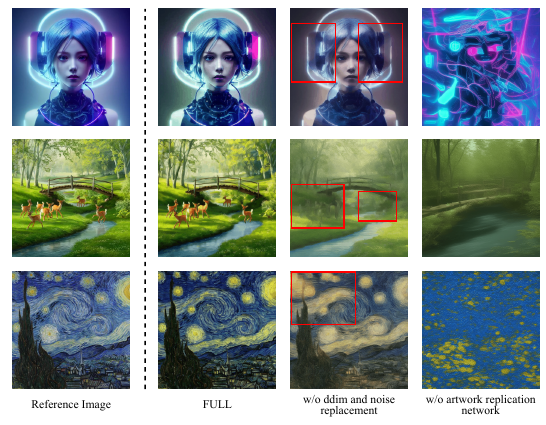}
    \caption{Ablation study on the consistency of the final frame reconstruction.}
    \label{fig:ablation}
\end{figure}


In this section, we conduct ablation studies to discuss the key designs in ProcessPainter that maintain consistency with the reference image. As shown in Figure \ref{fig:ablation}, when both the Artwork Replication Network and the noise replacement strategy are employed, the results are highly consistent with the reference image. When only the Artwork Replication Network is used to maintain consistency, the results have limited consistency with the reference image. When only the noise replacement strategy is used, the consistency with the reference image is poor.


\begin{table}[h!]
\centering
\footnotesize
\caption{User Preferences for Human-Like Painting Styles Comparison}
\begin{tabular}{lccc}
\toprule
Comparison Method & Anthropomorphic (\%) & Preference (\%)  \\
\midrule
Ours vs. LearnToPaint\cite{huang2019learning} & 78.2  & 68.2   \\
Ours vs. Paint Transformer\cite{liu2021paint} & 80.4  & 65.9   \\
Ours vs. Intelli-Paint\cite{singh2022intelli}  & 84.5 & 78.4   \\
Ours vs. Stylized Neural Painting\cite{snp}  & 78.6  & 71.6   \\
\bottomrule
\end{tabular}
\label{tab2}
\end{table}

\subsection{User Study}
The user study was conducted with 44 participants, each shown 28 paired painting sequences comparing our method to previous works. For each pair, participants were asked to select the painting sequence that best resembles the human painting style and which they preferred. Each sequence was presented as a GIF image with a total duration of 5 seconds. The results of the user study, shown in Table~\ref{tab2}, suggest that our outcomes more closely resemble the painting processes of human artists and are more favored.

\section{Limitation}
The training sequences of paintings in this paper use 8 frames with a resolution of 512x512. Due to GPU memory constraints, it is difficult to generate sequences with more frames or higher resolution. Another limitation is that, although many digital painting tools can export the creation process of artists, obtaining the painting processes of human artists, compared to static images, remains challenging.

\section{Conclusion}


In this paper, we introduce ProcessPainter, a novel framework that bridges the gap between digital image synthesis and traditional artistic practices by generating detailed painting processes from text descriptions. By training temporal models on both synthetic datasets and artists' painting sequence, ProcessPainter not only replicates the final visual styles of artists but also their unique creative processes. This approach provides a granular view of the artistic creation 'how-to' that is often missing in traditional diffusion model applications. We proposed the Artwork Replication Network, which enables controllable painting process generation, allowing for the conversion of reference images into process key-frames and the completion of semi-finished paintings. Our contributions to the field are substantial, providing a new methodology for art creation, analysis, and education that promises to enrich both the understanding and practice of digital and traditional arts.

\newpage
\bibliographystyle{ACM-Reference-Format}
\bibliography{acmart}


\begin{thebibliography}{67}


\ifx \showCODEN    \undefined \def \showCODEN     #1{\unskip}     \fi
\ifx \showDOI      \undefined \def \showDOI       #1{#1}\fi
\ifx \showISBNx    \undefined \def \showISBNx     #1{\unskip}     \fi
\ifx \showISBNxiii \undefined \def \showISBNxiii  #1{\unskip}     \fi
\ifx \showISSN     \undefined \def \showISSN      #1{\unskip}     \fi
\ifx \showLCCN     \undefined \def \showLCCN      #1{\unskip}     \fi
\ifx \shownote     \undefined \def \shownote      #1{#1}          \fi
\ifx \showarticletitle \undefined \def \showarticletitle #1{#1}   \fi
\ifx \showURL      \undefined \def \showURL       {\relax}        \fi
\providecommand\bibfield[2]{#2}
\providecommand\bibinfo[2]{#2}
\providecommand\natexlab[1]{#1}
\providecommand\showeprint[2][]{arXiv:#2}

\bibitem[Aksan et~al\mbox{.}(2020)]%
        {aksan2020cose}
\bibfield{author}{\bibinfo{person}{Emre Aksan}, \bibinfo{person}{Thomas Deselaers}, \bibinfo{person}{Andrea Tagliasacchi}, {and} \bibinfo{person}{Otmar Hilliges}.} \bibinfo{year}{2020}\natexlab{}.
\newblock \showarticletitle{Cose: Compositional stroke embeddings}.
\newblock \bibinfo{journal}{\emph{Advances in Neural Information Processing Systems}}  \bibinfo{volume}{33} (\bibinfo{year}{2020}), \bibinfo{pages}{10041--10052}.
\newblock


\bibitem[Blattmann et~al\mbox{.}(2023)]%
        {blattmann2023stable}
\bibfield{author}{\bibinfo{person}{Andreas Blattmann}, \bibinfo{person}{Tim Dockhorn}, \bibinfo{person}{Sumith Kulal}, \bibinfo{person}{Daniel Mendelevitch}, \bibinfo{person}{Maciej Kilian}, \bibinfo{person}{Dominik Lorenz}, \bibinfo{person}{Yam Levi}, \bibinfo{person}{Zion English}, \bibinfo{person}{Vikram Voleti}, \bibinfo{person}{Adam Letts}, {et~al\mbox{.}}} \bibinfo{year}{2023}\natexlab{}.
\newblock \showarticletitle{Stable video diffusion: Scaling latent video diffusion models to large datasets}.
\newblock \bibinfo{journal}{\emph{arXiv preprint arXiv:2311.15127}} (\bibinfo{year}{2023}).
\newblock


\bibitem[Cao et~al\mbox{.}(2019)]%
        {cao2019ai}
\bibfield{author}{\bibinfo{person}{Nan Cao}, \bibinfo{person}{Xin Yan}, \bibinfo{person}{Yang Shi}, {and} \bibinfo{person}{Chaoran Chen}.} \bibinfo{year}{2019}\natexlab{}.
\newblock \showarticletitle{AI-sketcher: a deep generative model for producing high-quality sketches}. In \bibinfo{booktitle}{\emph{Proceedings of the AAAI conference on artificial intelligence}}, Vol.~\bibinfo{volume}{33}. \bibinfo{pages}{2564--2571}.
\newblock


\bibitem[Cho et~al\mbox{.}(2024)]%
        {cho2024visual}
\bibfield{author}{\bibinfo{person}{Jaemin Cho}, \bibinfo{person}{Abhay Zala}, {and} \bibinfo{person}{Mohit Bansal}.} \bibinfo{year}{2024}\natexlab{}.
\newblock \showarticletitle{Visual Programming for Step-by-Step Text-to-Image Generation and Evaluation}.
\newblock \bibinfo{journal}{\emph{Advances in Neural Information Processing Systems}}  \bibinfo{volume}{36} (\bibinfo{year}{2024}).
\newblock


\bibitem[Dai et~al\mbox{.}(2023)]%
        {dai2023animateanything}
\bibfield{author}{\bibinfo{person}{Zuozhuo Dai}, \bibinfo{person}{Zhenghao Zhang}, \bibinfo{person}{Yao Yao}, \bibinfo{person}{Bingxue Qiu}, \bibinfo{person}{Siyu Zhu}, \bibinfo{person}{Long Qin}, {and} \bibinfo{person}{Weizhi Wang}.} \bibinfo{year}{2023}\natexlab{}.
\newblock \showarticletitle{AnimateAnything: Fine-Grained Open Domain Image Animation with Motion Guidance}.
\newblock \bibinfo{journal}{\emph{arXiv e-prints}} (\bibinfo{year}{2023}), \bibinfo{pages}{arXiv--2311}.
\newblock


\bibitem[Devlin et~al\mbox{.}(2018)]%
        {devlin2018bert}
\bibfield{author}{\bibinfo{person}{Jacob Devlin}, \bibinfo{person}{Ming-Wei Chang}, \bibinfo{person}{Kenton Lee}, {and} \bibinfo{person}{Kristina Toutanova}.} \bibinfo{year}{2018}\natexlab{}.
\newblock \showarticletitle{Bert: Pre-training of deep bidirectional transformers for language understanding}.
\newblock \bibinfo{journal}{\emph{arXiv preprint arXiv:1810.04805}} (\bibinfo{year}{2018}).
\newblock


\bibitem[Feng et~al\mbox{.}(2023)]%
        {feng2023ranni}
\bibfield{author}{\bibinfo{person}{Yutong Feng}, \bibinfo{person}{Biao Gong}, \bibinfo{person}{Di Chen}, \bibinfo{person}{Yujun Shen}, \bibinfo{person}{Yu Liu}, {and} \bibinfo{person}{Jingren Zhou}.} \bibinfo{year}{2023}\natexlab{}.
\newblock \showarticletitle{Ranni: Taming Text-to-Image Diffusion for Accurate Instruction Following}.
\newblock \bibinfo{journal}{\emph{arXiv preprint arXiv:2311.17002}} (\bibinfo{year}{2023}).
\newblock


\bibitem[Frans et~al\mbox{.}(2022)]%
        {clipdraw}
\bibfield{author}{\bibinfo{person}{Kevin Frans}, \bibinfo{person}{Lisa Soros}, {and} \bibinfo{person}{Olaf Witkowski}.} \bibinfo{year}{2022}\natexlab{}.
\newblock \showarticletitle{Clipdraw: Exploring text-to-drawing synthesis through language-image encoders}.
\newblock \bibinfo{journal}{\emph{Advances in Neural Information Processing Systems}}  \bibinfo{volume}{35} (\bibinfo{year}{2022}), \bibinfo{pages}{5207--5218}.
\newblock


\bibitem[Ge et~al\mbox{.}(2022)]%
        {ge2022long}
\bibfield{author}{\bibinfo{person}{Songwei Ge}, \bibinfo{person}{Thomas Hayes}, \bibinfo{person}{Harry Yang}, \bibinfo{person}{Xi Yin}, \bibinfo{person}{Guan Pang}, \bibinfo{person}{David Jacobs}, \bibinfo{person}{Jia-Bin Huang}, {and} \bibinfo{person}{Devi Parikh}.} \bibinfo{year}{2022}\natexlab{}.
\newblock \showarticletitle{Long video generation with time-agnostic vqgan and time-sensitive transformer}. In \bibinfo{booktitle}{\emph{European Conference on Computer Vision}}. Springer, \bibinfo{pages}{102--118}.
\newblock


\bibitem[Ge et~al\mbox{.}(2023)]%
        {ge2023preserve}
\bibfield{author}{\bibinfo{person}{Songwei Ge}, \bibinfo{person}{Seungjun Nah}, \bibinfo{person}{Guilin Liu}, \bibinfo{person}{Tyler Poon}, \bibinfo{person}{Andrew Tao}, \bibinfo{person}{Bryan Catanzaro}, \bibinfo{person}{David Jacobs}, \bibinfo{person}{Jia-Bin Huang}, \bibinfo{person}{Ming-Yu Liu}, {and} \bibinfo{person}{Yogesh Balaji}.} \bibinfo{year}{2023}\natexlab{}.
\newblock \showarticletitle{Preserve your own correlation: A noise prior for video diffusion models}. In \bibinfo{booktitle}{\emph{Proceedings of the IEEE/CVF International Conference on Computer Vision}}. \bibinfo{pages}{22930--22941}.
\newblock


\bibitem[Guo et~al\mbox{.}(2023)]%
        {guo2023animatediff}
\bibfield{author}{\bibinfo{person}{Yuwei Guo}, \bibinfo{person}{Ceyuan Yang}, \bibinfo{person}{Anyi Rao}, \bibinfo{person}{Yaohui Wang}, \bibinfo{person}{Yu Qiao}, \bibinfo{person}{Dahua Lin}, {and} \bibinfo{person}{Bo Dai}.} \bibinfo{year}{2023}\natexlab{}.
\newblock \showarticletitle{Animatediff: Animate your personalized text-to-image diffusion models without specific tuning}.
\newblock \bibinfo{journal}{\emph{arXiv preprint arXiv:2307.04725}} (\bibinfo{year}{2023}).
\newblock


\bibitem[Ha and Eck(2017)]%
        {ha2017neural}
\bibfield{author}{\bibinfo{person}{David Ha} {and} \bibinfo{person}{Douglas Eck}.} \bibinfo{year}{2017}\natexlab{}.
\newblock \showarticletitle{A neural representation of sketch drawings}.
\newblock \bibinfo{journal}{\emph{arXiv preprint arXiv:1704.03477}} (\bibinfo{year}{2017}).
\newblock


\bibitem[Haeberli(1990)]%
        {haeberli1990paint}
\bibfield{author}{\bibinfo{person}{Paul Haeberli}.} \bibinfo{year}{1990}\natexlab{}.
\newblock \showarticletitle{Paint by numbers: Abstract image representations}. In \bibinfo{booktitle}{\emph{Proceedings of the 17th annual conference on Computer graphics and interactive techniques}}. \bibinfo{pages}{207--214}.
\newblock


\bibitem[Hertzmann(2003)]%
        {hertzmann2003survey}
\bibfield{author}{\bibinfo{person}{Aaron Hertzmann}.} \bibinfo{year}{2003}\natexlab{}.
\newblock \showarticletitle{A survey of stroke-based rendering}. Institute of Electrical and Electronics Engineers.
\newblock


\bibitem[Hertzmann(2022)]%
        {hertzmann2022toward}
\bibfield{author}{\bibinfo{person}{Aaron Hertzmann}.} \bibinfo{year}{2022}\natexlab{}.
\newblock \showarticletitle{Toward modeling creative processes for algorithmic painting}.
\newblock \bibinfo{journal}{\emph{arXiv preprint arXiv:2205.01605}} (\bibinfo{year}{2022}).
\newblock


\bibitem[Ho et~al\mbox{.}(2020)]%
        {ho2020denoising}
\bibfield{author}{\bibinfo{person}{Jonathan Ho}, \bibinfo{person}{Ajay Jain}, {and} \bibinfo{person}{Pieter Abbeel}.} \bibinfo{year}{2020}\natexlab{}.
\newblock \showarticletitle{Denoising diffusion probabilistic models}.
\newblock \bibinfo{journal}{\emph{Advances in neural information processing systems}}  \bibinfo{volume}{33} (\bibinfo{year}{2020}), \bibinfo{pages}{6840--6851}.
\newblock


\bibitem[Hong et~al\mbox{.}(2022)]%
        {hong2022cogvideo}
\bibfield{author}{\bibinfo{person}{Wenyi Hong}, \bibinfo{person}{Ming Ding}, \bibinfo{person}{Wendi Zheng}, \bibinfo{person}{Xinghan Liu}, {and} \bibinfo{person}{Jie Tang}.} \bibinfo{year}{2022}\natexlab{}.
\newblock \showarticletitle{Cogvideo: Large-scale pretraining for text-to-video generation via transformers}.
\newblock \bibinfo{journal}{\emph{arXiv preprint arXiv:2205.15868}} (\bibinfo{year}{2022}).
\newblock


\bibitem[Hu et~al\mbox{.}(2021)]%
        {hu2021LoRA}
\bibfield{author}{\bibinfo{person}{Edward~J Hu}, \bibinfo{person}{Yelong Shen}, \bibinfo{person}{Phillip Wallis}, \bibinfo{person}{Zeyuan Allen-Zhu}, \bibinfo{person}{Yuanzhi Li}, \bibinfo{person}{Shean Wang}, \bibinfo{person}{Lu Wang}, {and} \bibinfo{person}{Weizhu Chen}.} \bibinfo{year}{2021}\natexlab{}.
\newblock \showarticletitle{Lora: Low-rank adaptation of large language models}.
\newblock \bibinfo{journal}{\emph{arXiv preprint arXiv:2106.09685}} (\bibinfo{year}{2021}).
\newblock


\bibitem[Hu et~al\mbox{.}(2023)]%
        {hu2023stroke}
\bibfield{author}{\bibinfo{person}{Teng Hu}, \bibinfo{person}{Ran Yi}, \bibinfo{person}{Haokun Zhu}, \bibinfo{person}{Liang Liu}, \bibinfo{person}{Jinlong Peng}, \bibinfo{person}{Yabiao Wang}, \bibinfo{person}{Chengjie Wang}, {and} \bibinfo{person}{Lizhuang Ma}.} \bibinfo{year}{2023}\natexlab{}.
\newblock \showarticletitle{Stroke-based Neural Painting and Stylization with Dynamically Predicted Painting Region}. In \bibinfo{booktitle}{\emph{Proceedings of the 31st ACM International Conference on Multimedia}}. \bibinfo{pages}{7470--7480}.
\newblock


\bibitem[Huang et~al\mbox{.}(2019)]%
        {huang2019learning}
\bibfield{author}{\bibinfo{person}{Zhewei Huang}, \bibinfo{person}{Wen Heng}, {and} \bibinfo{person}{Shuchang Zhou}.} \bibinfo{year}{2019}\natexlab{}.
\newblock \showarticletitle{Learning to paint with model-based deep reinforcement learning}. In \bibinfo{booktitle}{\emph{Proceedings of the IEEE/CVF international conference on computer vision}}. \bibinfo{pages}{8709--8718}.
\newblock


\bibitem[Khachatryan et~al\mbox{.}(2023)]%
        {khachatryan2023text2video}
\bibfield{author}{\bibinfo{person}{Levon Khachatryan}, \bibinfo{person}{Andranik Movsisyan}, \bibinfo{person}{Vahram Tadevosyan}, \bibinfo{person}{Roberto Henschel}, \bibinfo{person}{Zhangyang Wang}, \bibinfo{person}{Shant Navasardyan}, {and} \bibinfo{person}{Humphrey Shi}.} \bibinfo{year}{2023}\natexlab{}.
\newblock \showarticletitle{Text2video-zero: Text-to-image diffusion models are zero-shot video generators}. In \bibinfo{booktitle}{\emph{Proceedings of the IEEE/CVF International Conference on Computer Vision}}. \bibinfo{pages}{15954--15964}.
\newblock


\bibitem[Kirillov et~al\mbox{.}(2023)]%
        {kirillov2023segany}
\bibfield{author}{\bibinfo{person}{Alexander Kirillov}, \bibinfo{person}{Eric Mintun}, \bibinfo{person}{Nikhila Ravi}, \bibinfo{person}{Hanzi Mao}, \bibinfo{person}{Chloe Rolland}, \bibinfo{person}{Laura Gustafson}, \bibinfo{person}{Tete Xiao}, \bibinfo{person}{Spencer Whitehead}, \bibinfo{person}{Alexander~C. Berg}, \bibinfo{person}{Wan-Yen Lo}, \bibinfo{person}{Piotr Doll{\'a}r}, {and} \bibinfo{person}{Ross Girshick}.} \bibinfo{year}{2023}\natexlab{}.
\newblock \showarticletitle{Segment Anything}.
\newblock \bibinfo{journal}{\emph{arXiv:2304.02643}} (\bibinfo{year}{2023}).
\newblock


\bibitem[Kotovenko et~al\mbox{.}(2021)]%
        {kotovenko2021rethinking}
\bibfield{author}{\bibinfo{person}{Dmytro Kotovenko}, \bibinfo{person}{Matthias Wright}, \bibinfo{person}{Arthur Heimbrecht}, {and} \bibinfo{person}{Bjorn Ommer}.} \bibinfo{year}{2021}\natexlab{}.
\newblock \showarticletitle{Rethinking style transfer: From pixels to parameterized brushstrokes}. In \bibinfo{booktitle}{\emph{Proceedings of the IEEE/CVF Conference on Computer Vision and Pattern Recognition}}. \bibinfo{pages}{12196--12205}.
\newblock


\bibitem[Kumari et~al\mbox{.}(2023)]%
        {kumari2023multi}
\bibfield{author}{\bibinfo{person}{Nupur Kumari}, \bibinfo{person}{Bingliang Zhang}, \bibinfo{person}{Richard Zhang}, \bibinfo{person}{Eli Shechtman}, {and} \bibinfo{person}{Jun-Yan Zhu}.} \bibinfo{year}{2023}\natexlab{}.
\newblock \showarticletitle{Multi-concept customization of text-to-image diffusion}. In \bibinfo{booktitle}{\emph{Proceedings of the IEEE/CVF Conference on Computer Vision and Pattern Recognition}}. \bibinfo{pages}{1931--1941}.
\newblock


\bibitem[Le~Moing et~al\mbox{.}(2021)]%
        {le2021ccvs}
\bibfield{author}{\bibinfo{person}{Guillaume Le~Moing}, \bibinfo{person}{Jean Ponce}, {and} \bibinfo{person}{Cordelia Schmid}.} \bibinfo{year}{2021}\natexlab{}.
\newblock \showarticletitle{Ccvs: Context-aware controllable video synthesis}.
\newblock \bibinfo{journal}{\emph{Advances in Neural Information Processing Systems}}  \bibinfo{volume}{34} (\bibinfo{year}{2021}), \bibinfo{pages}{14042--14055}.
\newblock


\bibitem[Li et~al\mbox{.}(2024)]%
        {li2024blip}
\bibfield{author}{\bibinfo{person}{Dongxu Li}, \bibinfo{person}{Junnan Li}, {and} \bibinfo{person}{Steven Hoi}.} \bibinfo{year}{2024}\natexlab{}.
\newblock \showarticletitle{Blip-diffusion: Pre-trained subject representation for controllable text-to-image generation and editing}.
\newblock \bibinfo{journal}{\emph{Advances in Neural Information Processing Systems}}  \bibinfo{volume}{36} (\bibinfo{year}{2024}).
\newblock


\bibitem[Li et~al\mbox{.}(2023)]%
        {li2023gligen}
\bibfield{author}{\bibinfo{person}{Yuheng Li}, \bibinfo{person}{Haotian Liu}, \bibinfo{person}{Qingyang Wu}, \bibinfo{person}{Fangzhou Mu}, \bibinfo{person}{Jianwei Yang}, \bibinfo{person}{Jianfeng Gao}, \bibinfo{person}{Chunyuan Li}, {and} \bibinfo{person}{Yong~Jae Lee}.} \bibinfo{year}{2023}\natexlab{}.
\newblock \showarticletitle{Gligen: Open-set grounded text-to-image generation}. In \bibinfo{booktitle}{\emph{Proceedings of the IEEE/CVF Conference on Computer Vision and Pattern Recognition}}. \bibinfo{pages}{22511--22521}.
\newblock


\bibitem[Lin and Yang(2024)]%
        {lin2024animatediff}
\bibfield{author}{\bibinfo{person}{Shanchuan Lin} {and} \bibinfo{person}{Xiao Yang}.} \bibinfo{year}{2024}\natexlab{}.
\newblock \showarticletitle{AnimateDiff-Lightning: Cross-Model Diffusion Distillation}.
\newblock \bibinfo{journal}{\emph{arXiv preprint arXiv:2403.12706}} (\bibinfo{year}{2024}).
\newblock


\bibitem[Litwinowicz(1997)]%
        {litwinowicz1997processing}
\bibfield{author}{\bibinfo{person}{Peter Litwinowicz}.} \bibinfo{year}{1997}\natexlab{}.
\newblock \showarticletitle{Processing images and video for an impressionist effect}. In \bibinfo{booktitle}{\emph{Proceedings of the 24th annual conference on Computer graphics and interactive techniques}}. \bibinfo{pages}{407--414}.
\newblock


\bibitem[Liu et~al\mbox{.}(2021)]%
        {liu2021paint}
\bibfield{author}{\bibinfo{person}{Songhua Liu}, \bibinfo{person}{Tianwei Lin}, \bibinfo{person}{Dongliang He}, \bibinfo{person}{Fu Li}, \bibinfo{person}{Ruifeng Deng}, \bibinfo{person}{Xin Li}, \bibinfo{person}{Errui Ding}, {and} \bibinfo{person}{Hao Wang}.} \bibinfo{year}{2021}\natexlab{}.
\newblock \showarticletitle{Paint transformer: Feed forward neural painting with stroke prediction}. In \bibinfo{booktitle}{\emph{Proceedings of the IEEE/CVF international conference on computer vision}}. \bibinfo{pages}{6598--6607}.
\newblock


\bibitem[Ma et~al\mbox{.}(2024a)]%
        {ma2024follow2}
\bibfield{author}{\bibinfo{person}{Yue Ma}, \bibinfo{person}{Yingqing He}, \bibinfo{person}{Xiaodong Cun}, \bibinfo{person}{Xintao Wang}, \bibinfo{person}{Siran Chen}, \bibinfo{person}{Xiu Li}, {and} \bibinfo{person}{Qifeng Chen}.} \bibinfo{year}{2024}\natexlab{a}.
\newblock \showarticletitle{Follow your pose: Pose-guided text-to-video generation using pose-free videos}. In \bibinfo{booktitle}{\emph{Proceedings of the AAAI Conference on Artificial Intelligence}}, Vol.~\bibinfo{volume}{38}. \bibinfo{pages}{4117--4125}.
\newblock


\bibitem[Ma et~al\mbox{.}(2024b)]%
        {ma2024follow}
\bibfield{author}{\bibinfo{person}{Yue Ma}, \bibinfo{person}{Yingqing He}, \bibinfo{person}{Hongfa Wang}, \bibinfo{person}{Andong Wang}, \bibinfo{person}{Chenyang Qi}, \bibinfo{person}{Chengfei Cai}, \bibinfo{person}{Xiu Li}, \bibinfo{person}{Zhifeng Li}, \bibinfo{person}{Heung-Yeung Shum}, \bibinfo{person}{Wei Liu}, {et~al\mbox{.}}} \bibinfo{year}{2024}\natexlab{b}.
\newblock \showarticletitle{Follow-Your-Click: Open-domain Regional Image Animation via Short Prompts}.
\newblock \bibinfo{journal}{\emph{arXiv preprint arXiv:2403.08268}} (\bibinfo{year}{2024}).
\newblock


\bibitem[Nagahara(2023)]%
        {sparse}
\bibfield{author}{\bibinfo{person}{Masaaki Nagahara}.} \bibinfo{year}{2023}\natexlab{}.
\newblock \showarticletitle{Sparse control for continuous-time systems}.
\newblock \bibinfo{journal}{\emph{International Journal of Robust and Nonlinear Control}} \bibinfo{volume}{33}, \bibinfo{number}{1} (\bibinfo{year}{2023}), \bibinfo{pages}{6--22}.
\newblock


\bibitem[Nakano(2019)]%
        {nakano2019neural}
\bibfield{author}{\bibinfo{person}{Reiichiro Nakano}.} \bibinfo{year}{2019}\natexlab{}.
\newblock \showarticletitle{Neural painters: A learned differentiable constraint for generating brushstroke paintings}.
\newblock \bibinfo{journal}{\emph{arXiv preprint arXiv:1904.08410}} (\bibinfo{year}{2019}).
\newblock


\bibitem[Nichol et~al\mbox{.}(2021)]%
        {nichol2021glide}
\bibfield{author}{\bibinfo{person}{Alex Nichol}, \bibinfo{person}{Prafulla Dhariwal}, \bibinfo{person}{Aditya Ramesh}, \bibinfo{person}{Pranav Shyam}, \bibinfo{person}{Pamela Mishkin}, \bibinfo{person}{Bob McGrew}, \bibinfo{person}{Ilya Sutskever}, {and} \bibinfo{person}{Mark Chen}.} \bibinfo{year}{2021}\natexlab{}.
\newblock \showarticletitle{Glide: Towards photorealistic image generation and editing with text-guided diffusion models}.
\newblock \bibinfo{journal}{\emph{arXiv preprint arXiv:2112.10741}} (\bibinfo{year}{2021}).
\newblock


\bibitem[Podell et~al\mbox{.}(2023)]%
        {podell2023sdxl}
\bibfield{author}{\bibinfo{person}{Dustin Podell}, \bibinfo{person}{Zion English}, \bibinfo{person}{Kyle Lacey}, \bibinfo{person}{Andreas Blattmann}, \bibinfo{person}{Tim Dockhorn}, \bibinfo{person}{Jonas M{\"u}ller}, \bibinfo{person}{Joe Penna}, {and} \bibinfo{person}{Robin Rombach}.} \bibinfo{year}{2023}\natexlab{}.
\newblock \showarticletitle{Sdxl: Improving latent diffusion models for high-resolution image synthesis}.
\newblock \bibinfo{journal}{\emph{arXiv preprint arXiv:2307.01952}} (\bibinfo{year}{2023}).
\newblock


\bibitem[Radford et~al\mbox{.}(2021)]%
        {radford2021learning}
\bibfield{author}{\bibinfo{person}{Alec Radford}, \bibinfo{person}{Jong~Wook Kim}, \bibinfo{person}{Chris Hallacy}, \bibinfo{person}{Aditya Ramesh}, \bibinfo{person}{Gabriel Goh}, \bibinfo{person}{Sandhini Agarwal}, \bibinfo{person}{Girish Sastry}, \bibinfo{person}{Amanda Askell}, \bibinfo{person}{Pamela Mishkin}, \bibinfo{person}{Jack Clark}, {et~al\mbox{.}}} \bibinfo{year}{2021}\natexlab{}.
\newblock \showarticletitle{Learning transferable visual models from natural language supervision}. In \bibinfo{booktitle}{\emph{International conference on machine learning}}. PMLR, \bibinfo{pages}{8748--8763}.
\newblock


\bibitem[Ramesh et~al\mbox{.}(2022)]%
        {ramesh2022hierarchical}
\bibfield{author}{\bibinfo{person}{Aditya Ramesh}, \bibinfo{person}{Prafulla Dhariwal}, \bibinfo{person}{Alex Nichol}, \bibinfo{person}{Casey Chu}, {and} \bibinfo{person}{Mark Chen}.} \bibinfo{year}{2022}\natexlab{}.
\newblock \showarticletitle{Hierarchical text-conditional image generation with clip latents. arXiv 2022}.
\newblock \bibinfo{journal}{\emph{arXiv preprint arXiv:2204.06125}} (\bibinfo{year}{2022}).
\newblock


\bibitem[Rombach et~al\mbox{.}(2022)]%
        {rombach2022high}
\bibfield{author}{\bibinfo{person}{Robin Rombach}, \bibinfo{person}{Andreas Blattmann}, \bibinfo{person}{Dominik Lorenz}, \bibinfo{person}{Patrick Esser}, {and} \bibinfo{person}{Bj{\"o}rn Ommer}.} \bibinfo{year}{2022}\natexlab{}.
\newblock \showarticletitle{High-resolution image synthesis with latent diffusion models}. In \bibinfo{booktitle}{\emph{Proceedings of the IEEE/CVF conference on computer vision and pattern recognition}}. \bibinfo{pages}{10684--10695}.
\newblock


\bibitem[Ruiz et~al\mbox{.}(2023)]%
        {ruiz2023dreambooth}
\bibfield{author}{\bibinfo{person}{Nataniel Ruiz}, \bibinfo{person}{Yuanzhen Li}, \bibinfo{person}{Varun Jampani}, \bibinfo{person}{Yael Pritch}, \bibinfo{person}{Michael Rubinstein}, {and} \bibinfo{person}{Kfir Aberman}.} \bibinfo{year}{2023}\natexlab{}.
\newblock \showarticletitle{Dreambooth: Fine tuning text-to-image diffusion models for subject-driven generation}. In \bibinfo{booktitle}{\emph{Proceedings of the IEEE/CVF Conference on Computer Vision and Pattern Recognition}}. \bibinfo{pages}{22500--22510}.
\newblock


\bibitem[Saharia et~al\mbox{.}(2022)]%
        {saharia2022photorealistic}
\bibfield{author}{\bibinfo{person}{Chitwan Saharia}, \bibinfo{person}{William Chan}, \bibinfo{person}{Saurabh Saxena}, \bibinfo{person}{Lala Li}, \bibinfo{person}{Jay Whang}, \bibinfo{person}{Emily~L Denton}, \bibinfo{person}{Kamyar Ghasemipour}, \bibinfo{person}{Raphael Gontijo~Lopes}, \bibinfo{person}{Burcu Karagol~Ayan}, \bibinfo{person}{Tim Salimans}, {et~al\mbox{.}}} \bibinfo{year}{2022}\natexlab{}.
\newblock \showarticletitle{Photorealistic text-to-image diffusion models with deep language understanding}.
\newblock \bibinfo{journal}{\emph{Advances in neural information processing systems}}  \bibinfo{volume}{35} (\bibinfo{year}{2022}), \bibinfo{pages}{36479--36494}.
\newblock


\bibitem[Schaldenbrand and Oh(2021)]%
        {schaldenbrand2021content}
\bibfield{author}{\bibinfo{person}{Peter Schaldenbrand} {and} \bibinfo{person}{Jean Oh}.} \bibinfo{year}{2021}\natexlab{}.
\newblock \showarticletitle{Content masked loss: Human-like brush stroke planning in a reinforcement learning painting agent}. In \bibinfo{booktitle}{\emph{Proceedings of the AAAI conference on artificial intelligence}}, Vol.~\bibinfo{volume}{35}. \bibinfo{pages}{505--512}.
\newblock


\bibitem[Schuhmann et~al\mbox{.}(2021)]%
        {schuhmann2021laion}
\bibfield{author}{\bibinfo{person}{Christoph Schuhmann}, \bibinfo{person}{Richard Vencu}, \bibinfo{person}{Romain Beaumont}, \bibinfo{person}{Robert Kaczmarczyk}, \bibinfo{person}{Clayton Mullis}, \bibinfo{person}{Aarush Katta}, \bibinfo{person}{Theo Coombes}, \bibinfo{person}{Jenia Jitsev}, {and} \bibinfo{person}{Aran Komatsuzaki}.} \bibinfo{year}{2021}\natexlab{}.
\newblock \showarticletitle{Laion-400m: Open dataset of clip-filtered 400 million image-text pairs}.
\newblock \bibinfo{journal}{\emph{arXiv preprint arXiv:2111.02114}} (\bibinfo{year}{2021}).
\newblock


\bibitem[Shen et~al\mbox{.}(2023)]%
        {shen2023mostgan}
\bibfield{author}{\bibinfo{person}{Xiaoqian Shen}, \bibinfo{person}{Xiang Li}, {and} \bibinfo{person}{Mohamed Elhoseiny}.} \bibinfo{year}{2023}\natexlab{}.
\newblock \showarticletitle{Mostgan-v: Video generation with temporal motion styles}. In \bibinfo{booktitle}{\emph{Proceedings of the IEEE/CVF Conference on Computer Vision and Pattern Recognition}}. \bibinfo{pages}{5652--5661}.
\newblock


\bibitem[Singh et~al\mbox{.}(2022)]%
        {singh2022intelli}
\bibfield{author}{\bibinfo{person}{Jaskirat Singh}, \bibinfo{person}{Cameron Smith}, \bibinfo{person}{Jose Echevarria}, {and} \bibinfo{person}{Liang Zheng}.} \bibinfo{year}{2022}\natexlab{}.
\newblock \showarticletitle{Intelli-Paint: Towards developing more human-intelligible painting agents}. In \bibinfo{booktitle}{\emph{European Conference on Computer Vision}}. Springer, \bibinfo{pages}{685--701}.
\newblock


\bibitem[Song et~al\mbox{.}(2020)]%
        {song2020denoising}
\bibfield{author}{\bibinfo{person}{Jiaming Song}, \bibinfo{person}{Chenlin Meng}, {and} \bibinfo{person}{Stefano Ermon}.} \bibinfo{year}{2020}\natexlab{}.
\newblock \showarticletitle{Denoising diffusion implicit models}.
\newblock \bibinfo{journal}{\emph{arXiv preprint arXiv:2010.02502}} (\bibinfo{year}{2020}).
\newblock


\bibitem[Song(2022)]%
        {cliptexture}
\bibfield{author}{\bibinfo{person}{Yiren Song}.} \bibinfo{year}{2022}\natexlab{}.
\newblock \showarticletitle{Cliptexture: Text-driven texture synthesis}. In \bibinfo{booktitle}{\emph{Proceedings of the 30th ACM International Conference on Multimedia}}. \bibinfo{pages}{5468--5476}.
\newblock


\bibitem[Song et~al\mbox{.}(2022)]%
        {3Dstroke}
\bibfield{author}{\bibinfo{person}{Yiren Song}, \bibinfo{person}{Zhongliang Jing}, {and} \bibinfo{person}{Minzhe Li}.} \bibinfo{year}{2022}\natexlab{}.
\newblock \showarticletitle{Stroke Based Painting with 3D Perception}. In \bibinfo{booktitle}{\emph{Chinese Conference on Image and Graphics Technologies}}. Springer, \bibinfo{pages}{326--341}.
\newblock


\bibitem[Song et~al\mbox{.}(2023)]%
        {clipvg}
\bibfield{author}{\bibinfo{person}{Yiren Song}, \bibinfo{person}{Xuning Shao}, \bibinfo{person}{Kang Chen}, \bibinfo{person}{Weidong Zhang}, \bibinfo{person}{Zhongliang Jing}, {and} \bibinfo{person}{Minzhe Li}.} \bibinfo{year}{2023}\natexlab{}.
\newblock \showarticletitle{Clipvg: Text-guided image manipulation using differentiable vector graphics}. In \bibinfo{booktitle}{\emph{Proceedings of the AAAI Conference on Artificial Intelligence}}, Vol.~\bibinfo{volume}{37}. \bibinfo{pages}{2312--2320}.
\newblock


\bibitem[Song and Zhang(2022)]%
        {clipfont}
\bibfield{author}{\bibinfo{person}{Yiren Song} {and} \bibinfo{person}{Yuxuan Zhang}.} \bibinfo{year}{2022}\natexlab{}.
\newblock \showarticletitle{CLIPFont: Text Guided Vector WordArt Generation.}. In \bibinfo{booktitle}{\emph{BMVC}}. \bibinfo{pages}{543}.
\newblock


\bibitem[Tian et~al\mbox{.}(2021)]%
        {tian2021good}
\bibfield{author}{\bibinfo{person}{Yu Tian}, \bibinfo{person}{Jian Ren}, \bibinfo{person}{Menglei Chai}, \bibinfo{person}{Kyle Olszewski}, \bibinfo{person}{Xi Peng}, \bibinfo{person}{Dimitris~N Metaxas}, {and} \bibinfo{person}{Sergey Tulyakov}.} \bibinfo{year}{2021}\natexlab{}.
\newblock \showarticletitle{A good image generator is what you need for high-resolution video synthesis}.
\newblock \bibinfo{journal}{\emph{arXiv preprint arXiv:2104.15069}} (\bibinfo{year}{2021}).
\newblock


\bibitem[Tong et~al\mbox{.}(2022)]%
        {tong2022im2oil}
\bibfield{author}{\bibinfo{person}{Zhengyan Tong}, \bibinfo{person}{Xiaohang Wang}, \bibinfo{person}{Shengchao Yuan}, \bibinfo{person}{Xuanhong Chen}, \bibinfo{person}{Junjie Wang}, {and} \bibinfo{person}{Xiangzhong Fang}.} \bibinfo{year}{2022}\natexlab{}.
\newblock \showarticletitle{Im2oil: Stroke-based oil painting rendering with linearly controllable fineness via adaptive sampling}. In \bibinfo{booktitle}{\emph{Proceedings of the 30th ACM International Conference on Multimedia}}. \bibinfo{pages}{1035--1046}.
\newblock


\bibitem[Vanderhaeghe and Collomosse(2012)]%
        {vanderhaeghe2012stroke}
\bibfield{author}{\bibinfo{person}{David Vanderhaeghe} {and} \bibinfo{person}{John Collomosse}.} \bibinfo{year}{2012}\natexlab{}.
\newblock \showarticletitle{Stroke based painterly rendering}.
\newblock In \bibinfo{booktitle}{\emph{Image and Video-Based Artistic Stylisation}}. \bibinfo{publisher}{Springer}, \bibinfo{pages}{3--21}.
\newblock


\bibitem[Wang et~al\mbox{.}(2024a)]%
        {wang2024instantid}
\bibfield{author}{\bibinfo{person}{Qixun Wang}, \bibinfo{person}{Xu Bai}, \bibinfo{person}{Haofan Wang}, \bibinfo{person}{Zekui Qin}, {and} \bibinfo{person}{Anthony Chen}.} \bibinfo{year}{2024}\natexlab{a}.
\newblock \showarticletitle{Instantid: Zero-shot identity-preserving generation in seconds}.
\newblock \bibinfo{journal}{\emph{arXiv preprint arXiv:2401.07519}} (\bibinfo{year}{2024}).
\newblock


\bibitem[Wang et~al\mbox{.}(2024b)]%
        {wang2024stablegarment}
\bibfield{author}{\bibinfo{person}{Rui Wang}, \bibinfo{person}{Hailong Guo}, \bibinfo{person}{Jiaming Liu}, \bibinfo{person}{Huaxia Li}, \bibinfo{person}{Haibo Zhao}, \bibinfo{person}{Xu Tang}, \bibinfo{person}{Yao Hu}, \bibinfo{person}{Hao Tang}, {and} \bibinfo{person}{Peipei Li}.} \bibinfo{year}{2024}\natexlab{b}.
\newblock \showarticletitle{StableGarment: Garment-Centric Generation via Stable Diffusion}.
\newblock \bibinfo{journal}{\emph{arXiv preprint arXiv:2403.10783}} (\bibinfo{year}{2024}).
\newblock


\bibitem[Wang et~al\mbox{.}(2023)]%
        {wang2023motionctrl}
\bibfield{author}{\bibinfo{person}{Zhouxia Wang}, \bibinfo{person}{Ziyang Yuan}, \bibinfo{person}{Xintao Wang}, \bibinfo{person}{Tianshui Chen}, \bibinfo{person}{Menghan Xia}, \bibinfo{person}{Ping Luo}, {and} \bibinfo{person}{Ying Shan}.} \bibinfo{year}{2023}\natexlab{}.
\newblock \showarticletitle{Motionctrl: A unified and flexible motion controller for video generation}.
\newblock \bibinfo{journal}{\emph{arXiv preprint arXiv:2312.03641}} (\bibinfo{year}{2023}).
\newblock


\bibitem[Xie et~al\mbox{.}(2013)]%
        {xie2013artist}
\bibfield{author}{\bibinfo{person}{Ning Xie}, \bibinfo{person}{Hirotaka Hachiya}, {and} \bibinfo{person}{Masashi Sugiyama}.} \bibinfo{year}{2013}\natexlab{}.
\newblock \showarticletitle{Artist agent: A reinforcement learning approach to automatic stroke generation in oriental ink painting}.
\newblock \bibinfo{journal}{\emph{IEICE TRANSACTIONS on Information and Systems}} \bibinfo{volume}{96}, \bibinfo{number}{5} (\bibinfo{year}{2013}), \bibinfo{pages}{1134--1144}.
\newblock


\bibitem[Yang et~al\mbox{.}(2024)]%
        {depthanything}
\bibfield{author}{\bibinfo{person}{Lihe Yang}, \bibinfo{person}{Bingyi Kang}, \bibinfo{person}{Zilong Huang}, \bibinfo{person}{Xiaogang Xu}, \bibinfo{person}{Jiashi Feng}, {and} \bibinfo{person}{Hengshuang Zhao}.} \bibinfo{year}{2024}\natexlab{}.
\newblock \showarticletitle{Depth Anything: Unleashing the Power of Large-Scale Unlabeled Data}. In \bibinfo{booktitle}{\emph{CVPR}}.
\newblock


\bibitem[Yang et~al\mbox{.}(2023)]%
        {yang2023reco}
\bibfield{author}{\bibinfo{person}{Zhengyuan Yang}, \bibinfo{person}{Jianfeng Wang}, \bibinfo{person}{Zhe Gan}, \bibinfo{person}{Linjie Li}, \bibinfo{person}{Kevin Lin}, \bibinfo{person}{Chenfei Wu}, \bibinfo{person}{Nan Duan}, \bibinfo{person}{Zicheng Liu}, \bibinfo{person}{Ce Liu}, \bibinfo{person}{Michael Zeng}, {et~al\mbox{.}}} \bibinfo{year}{2023}\natexlab{}.
\newblock \showarticletitle{Reco: Region-controlled text-to-image generation}. In \bibinfo{booktitle}{\emph{Proceedings of the IEEE/CVF Conference on Computer Vision and Pattern Recognition}}. \bibinfo{pages}{14246--14255}.
\newblock


\bibitem[Ye et~al\mbox{.}(2023)]%
        {ye2023ip}
\bibfield{author}{\bibinfo{person}{Hu Ye}, \bibinfo{person}{Jun Zhang}, \bibinfo{person}{Sibo Liu}, \bibinfo{person}{Xiao Han}, {and} \bibinfo{person}{Wei Yang}.} \bibinfo{year}{2023}\natexlab{}.
\newblock \showarticletitle{Ip-adapter: Text compatible image prompt adapter for text-to-image diffusion models}.
\newblock \bibinfo{journal}{\emph{arXiv preprint arXiv:2308.06721}} (\bibinfo{year}{2023}).
\newblock


\bibitem[Zhang et~al\mbox{.}(2023a)]%
        {zhang2023adding}
\bibfield{author}{\bibinfo{person}{Lvmin Zhang}, \bibinfo{person}{Anyi Rao}, {and} \bibinfo{person}{Maneesh Agrawala}.} \bibinfo{year}{2023}\natexlab{a}.
\newblock \showarticletitle{Adding conditional control to text-to-image diffusion models}. In \bibinfo{booktitle}{\emph{Proceedings of the IEEE/CVF International Conference on Computer Vision}}. \bibinfo{pages}{3836--3847}.
\newblock


\bibitem[Zhang et~al\mbox{.}(2024a)]%
        {ssr}
\bibfield{author}{\bibinfo{person}{Yuxuan Zhang}, \bibinfo{person}{Yiren Song}, \bibinfo{person}{Jiaming Liu}, \bibinfo{person}{Rui Wang}, \bibinfo{person}{Jinpeng Yu}, \bibinfo{person}{Hao Tang}, \bibinfo{person}{Huaxia Li}, \bibinfo{person}{Xu Tang}, \bibinfo{person}{Yao Hu}, \bibinfo{person}{Han Pan}, {and} \bibinfo{person}{Zhongliang Jing}.} \bibinfo{year}{2024}\natexlab{a}.
\newblock \showarticletitle{SSR-Encoder: Encoding Selective Subject Representation for Subject-Driven Generation}. In \bibinfo{booktitle}{\emph{Proceedings of the IEEE/CVF Conference on Computer Vision and Pattern Recognition (CVPR)}}. \bibinfo{pages}{8069--8078}.
\newblock


\bibitem[Zhang et~al\mbox{.}(2024b)]%
        {fast_icassp}
\bibfield{author}{\bibinfo{person}{Yuxuan Zhang}, \bibinfo{person}{Yiren Song}, \bibinfo{person}{Jinpeng Yu}, \bibinfo{person}{Han Pan}, {and} \bibinfo{person}{Zhongliang Jing}.} \bibinfo{year}{2024}\natexlab{b}.
\newblock \showarticletitle{Fast Personalized Text to Image Synthesis with Attention Injection}. In \bibinfo{booktitle}{\emph{ICASSP 2024 - 2024 IEEE International Conference on Acoustics, Speech and Signal Processing (ICASSP)}}. \bibinfo{pages}{6195--6199}.
\newblock
\urldef\tempurl%
\url{https://doi.org/10.1109/ICASSP48485.2024.10447042}
\showDOI{\tempurl}


\bibitem[Zhang et~al\mbox{.}(2023b)]%
        {zhang2023controlvideo}
\bibfield{author}{\bibinfo{person}{Yabo Zhang}, \bibinfo{person}{Yuxiang Wei}, \bibinfo{person}{Dongsheng Jiang}, \bibinfo{person}{Xiaopeng Zhang}, \bibinfo{person}{Wangmeng Zuo}, {and} \bibinfo{person}{Qi Tian}.} \bibinfo{year}{2023}\natexlab{b}.
\newblock \showarticletitle{Controlvideo: Training-free controllable text-to-video generation}.
\newblock \bibinfo{journal}{\emph{arXiv preprint arXiv:2305.13077}} (\bibinfo{year}{2023}).
\newblock


\bibitem[Zhou et~al\mbox{.}(2018)]%
        {zhou2018learning}
\bibfield{author}{\bibinfo{person}{Tao Zhou}, \bibinfo{person}{Chen Fang}, \bibinfo{person}{Zhaowen Wang}, \bibinfo{person}{Jimei Yang}, \bibinfo{person}{Byungmoon Kim}, \bibinfo{person}{Zhili Chen}, \bibinfo{person}{Jonathan Brandt}, {and} \bibinfo{person}{Demetri Terzopoulos}.} \bibinfo{year}{2018}\natexlab{}.
\newblock \showarticletitle{Learning to sketch with deep q networks and demonstrated strokes}.
\newblock \bibinfo{journal}{\emph{arXiv preprint arXiv:1810.05977}} (\bibinfo{year}{2018}).
\newblock


\bibitem[Zhou et~al\mbox{.}(2021)]%
        {zhou2021lafite}
\bibfield{author}{\bibinfo{person}{Y Zhou}, \bibinfo{person}{R Zhang}, \bibinfo{person}{C Chen}, \bibinfo{person}{C Li}, \bibinfo{person}{C Tensmeyer}, \bibinfo{person}{T Yu}, \bibinfo{person}{J Gu}, \bibinfo{person}{J Xu}, {and} \bibinfo{person}{T Sun}.} \bibinfo{year}{2021}\natexlab{}.
\newblock \showarticletitle{Lafite: Towards language-free training for text-to-image generation. arxiv 2021}.
\newblock \bibinfo{journal}{\emph{arXiv preprint arXiv:2111.13792}}  \bibinfo{volume}{2} (\bibinfo{year}{2021}).
\newblock


\bibitem[Zou et~al\mbox{.}(2021)]%
        {snp}
\bibfield{author}{\bibinfo{person}{Zhengxia Zou}, \bibinfo{person}{Tianyang Shi}, \bibinfo{person}{Shuang Qiu}, \bibinfo{person}{Yi Yuan}, {and} \bibinfo{person}{Zhenwei Shi}.} \bibinfo{year}{2021}\natexlab{}.
\newblock \showarticletitle{Stylized neural painting}. In \bibinfo{booktitle}{\emph{Proceedings of the IEEE/CVF Conference on Computer Vision and Pattern Recognition}}. \bibinfo{pages}{15689--15698}.
\newblock


\end{thebibliography}









\end{document}